\documentclass[11pt,a4paper]{article}
\usepackage[hyperref]{emnlp2018}
\usepackage{times}
\usepackage{latexsym}
\usepackage{graphicx}
\usepackage{amsmath}
\usepackage{amsfonts}
\usepackage{amssymb}
\usepackage{caption}
\usepackage{booktabs}

\usepackage{url}

\aclfinalcopy 

\title{Ranking Paragraphs for Improving Answer Recall in\\Open-Domain Question Answering}

\author{Jinhyuk Lee, Seongjun Yun, Hyunjae Kim, Miyoung Ko\and Jaewoo Kang\Thanks { Corresponding author} \\
  Department of Computer Science and Engineering, Korea University \\
  {\tt \{jinhyuk\_lee, ysj5419, hyunjae-kim\}@korea.ac.kr} \\
  {\tt \{gomi1503, kangj\}@korea.ac.kr} \\
}

\begin{document}
\maketitle
\begin{abstract}
Recently, open-domain question answering (QA) has been combined with machine comprehension models to find answers in a large knowledge source. As open-domain QA requires retrieving relevant documents from text corpora to answer questions, its performance largely depends on the performance of document retrievers. However, since traditional information retrieval systems are not effective in obtaining documents with a high probability of containing answers, they lower the performance of QA systems. Simply extracting more documents increases the number of irrelevant documents, which also degrades the performance of QA systems. In this paper, we introduce Paragraph Ranker which ranks paragraphs of retrieved documents for a higher answer recall with less noise. We show that ranking paragraphs and aggregating answers using Paragraph Ranker improves performance of open-domain QA pipeline on the four open-domain QA datasets by 7.8\% on average.
\end{abstract}

\section{Introduction}
With the introduction of large scale machine comprehension datasets, machine comprehension models that are highly accurate and efficient in answering questions given raw texts have been proposed recently \cite{seo2016bidirectional,xiong2016dynamic,wang2017gated}. While conventional machine comprehension models were given a paragraph that always contains an answer to a question, some researchers have extended the models to an open-domain setting where relevant documents have to be searched from an extremely large knowledge source such as Wikipedia \cite{chen2017reading,wang2017r}. However, most of the open-domain QA pipelines depend on traditional information retrieval systems which use TF-IDF rankings \cite{chen2017reading,wang2017evidence}. Despite the efficiency of the traditional retrieval systems, the documents retrieved and ranked at the top by such systems often do not contain answers to questions. However, simply increasing the number of top ranked documents to find answers also increases the number of irrelevant documents. The tradeoff between reading more documents and minimizing noise is frequently observed in previous works that defined the $N$ number of top documents as a hyper-parameter to find \cite{wang2017r}. \\
\indent In this paper, we tackle the problem of ranking the paragraphs of retrieved documents for improving the answer recall of the paragraphs while filtering irrelevant paragraphs. By using our simple but efficient Paragraph Ranker, our QA pipeline considers more documents for a high answer recall, and ranks paragraphs to read only the most relevant ones. The work closest to ours is that of Wang et al. \shortcite{wang2017r}. However, whereas their main focus is on re-ranking retrieved sentences to maximize the rewards of correctly answering the questions, our focus is to increase the answer recall of paragraphs with less noise. Thus, our work is complementary to the work of Wang et al. \shortcite{wang2017r}.

\begin{figure*}
\begin{center}
\includegraphics[width=15.2cm]{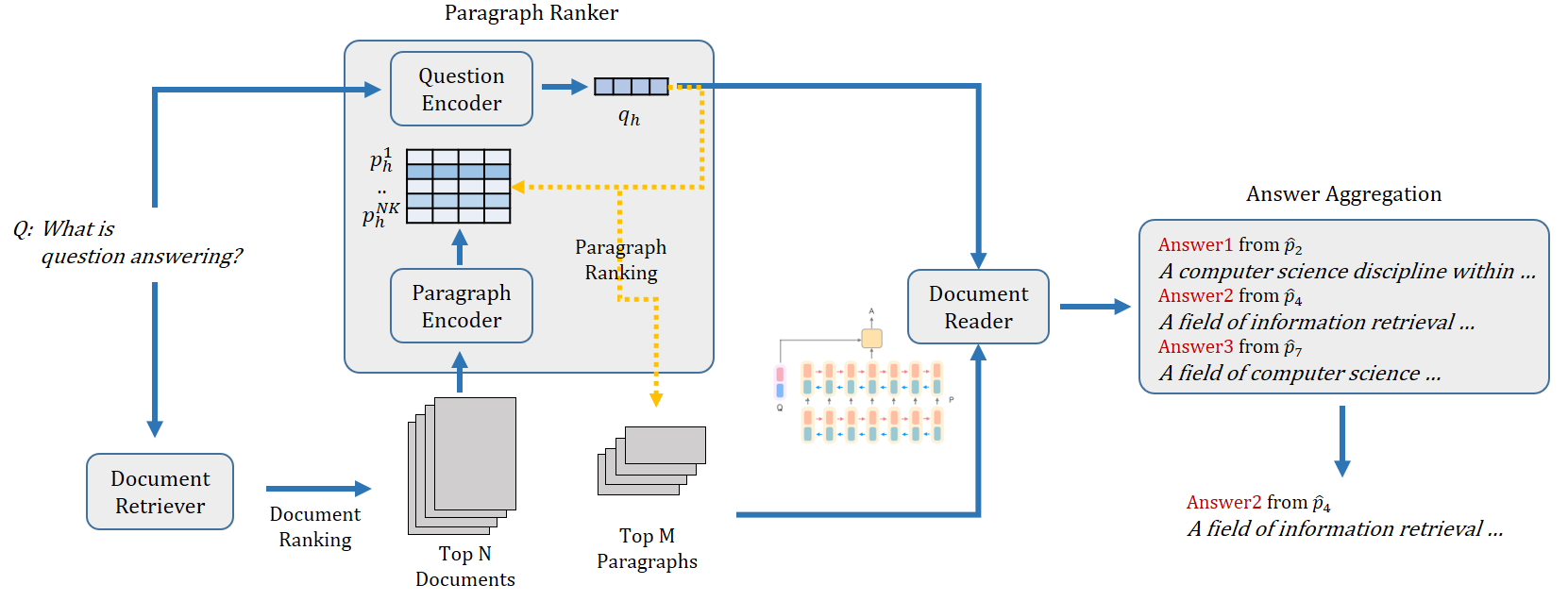}
\caption{Our proposed open-domain QA pipeline with Paragraph Ranker}
\label{fig:model_overview}
\end{center}
\end{figure*}

\indent Our work is largely inspired by the field of information retrieval called \textit{Learning to Rank} \cite{liu2009learning,severyn2015learning}. Most learning to rank models consist of two parts: encoding networks and ranking functions. We use bidirectional long short term memory (Bi-LSTM) as our encoding network, and apply various ranking functions proposed by previous works \cite{severyn2015learning,tu2017exploration}. Also, as the time and space complexities of ranking paragraphs are much larger than those of ranking sentences \cite{severyn2015learning}, we resort to negative sampling \cite{mikolov2013distributed} for an efficient training of our Paragraph Ranker. \\
\indent Our pipeline with Paragraph Ranker improves the exact match scores on the four open-domain QA datasets by 7.8\% on average. Even though we did not further customize Document Reader of DrQA \cite{chen2017reading}, the large improvement in the exact match scores shows that future researches would benefit from ranking and reading the more relevant paragraphs. By a qualitative analysis of ranked paragraphs, we provide additional evidence supporting our findings.

\section{Open-Domain QA Pipeline}
Most open-domain QA systems are constructed as pipelines that include a retrieval system and a reader model. We additionally built Paragraph Ranker that assists our QA pipeline for a better paragraph selection. For the retrieval system and the reader model, we used Document Retriever and Document Reader of Chen et al. \shortcite{chen2017reading}.\footnote{https://github.com/facebookresearch/DrQA} The overview of our pipeline is illustrated in Figure \ref{fig:model_overview}.

\subsection{Paragraph Ranker}
Given $N$ number of documents retrieved from Document Retriever, we assume that each document contains $K$ number of paragraphs on average. Instead of feeding all $NK$ number of paragraphs to Document Reader, we select only $M$ number of paragraphs using Paragraph Ranker. Utilizing Paragraph Ranker, we safely increase $N$ for a higher answer recall, and reduce the number of paragraphs to read by selecting only top ranked paragraphs. \\
\indent Given the retrieved paragraphs $P_i$ where $i$ ranges from $1$ to $NK$, and a question $Q$, we encode each paragraph and the question using two separate RNNs such as Bi-LSTM. Representations of each paragraph and the question are calculated as follows:
\begin{equation*}
p_h^i = \text{BiLSTM}_p(\text{E}(P_i)) \quad q_h = \text{BiLSTM}_q(\text{E}(Q))
\end{equation*}
\noindent where \text{BiLSTM($\cdot$)} returns the concatenation of the last hidden state of forward LSTM and the first hidden state of backward LSTM. \text{E}($\cdot$) converts tokens in a paragraph or a question into pretrained word embeddings. We use GloVe \cite{pennington2014glove} for the pretrained word embeddings.\\
\indent Once each paragraph and the question are represented as $p_h^i$ and $q_h$, we calculate the probability of each paragraph to contain an answer of the question as follows:
\begin{equation*}
p(P_i|Q) = \frac{1}{1+e^{-s(p_h^i, q_h)}}
\end{equation*}
\noindent where we have used similarity function $s(\cdot, \cdot)$ to measure the probability of containing answer to the question $Q$ in the paragraph $P_i$. While Wang and Jiang \shortcite{wang2015learning} adopted high capacity models such as Match-LSTM for measuring the similarity between paragraphs and questions, we use much simpler scoring functions to calculate the similarity more efficiently. We tested three different scoring functions: 1) the dot product of $p_h^i$ and $q_h$, 2) the bilinear form ${p_h^i}^TWq_h$, and 3) a multilayer perceptron (MLP) \cite{severyn2015learning}. While utilizing MLP takes much more time than the other two functions, recall of MLP was similar to that of the dot product. Also, as recall of the bilinear form was worse than that of the dot product, we use the dot product as our scoring function. \\
\indent Due to the large size of $NK$, it is difficult to train Paragraph Ranker on all the retrieved paragraphs.\footnote{$NK \approx 350$ when $N=5$ in SQuAD QA pairs.} To efficiently train our model, we use a negative sampling of irrelevant paragraphs \cite{
mikolov2013distributed}. Hence, the loss function of our model is as follows:
\begin{align*}
J(\Theta) = &-\log p(P_i|Q) \notag \\
		  & -E_{k\sim p_n}[\log (1-p(P_k|Q))] \notag
\end{align*}
\noindent where $k$ indicates indexes of negative samples that do not contain the answer, and $\Theta$ denotes trainable parameters of Paragraph Ranker. The distribution of negative samples are defined as $p_n$. We use the distribution of all the Stanford Question Answering Dataset (SQuAD) \cite{rajpurkar2016squad} training paragraphs as $p_n$ . \\
\indent Based on the rank of each paragraph from Paragraph Ranker and the rank of source document from Document Retriever, we collect top $M$ paragraphs to read. We combine the ranks by the multiplication of probabilities $p(P_i|Q)$ and $\tilde{p}(D_i|Q)$ to find most relevant paragraphs where $\tilde{p}(D_i|Q)$ denotes TF-IDF score of a source document $D_i$.

\begin{table*}[t]
\centering
\begin{tabular}{lcccccccc}
\toprule
& \multicolumn{2}{c}{$\text{SQuAD}_\text{OPEN}$} & \multicolumn{2}{c}{CuratedTrec} & \multicolumn{2}{c}{WebQuestions} & \multicolumn{2}{c}{WikiMovies} \\
\cmidrule{2-9}
Model & EM & Recall & EM & Recall & EM & Recall & EM & Recall \\ 
\midrule
DrQA \cite{chen2017reading} & 27.1 & 77.8 & 19.7 & 86.0 & 11.8 & 74.4 & 24.5 & 70.3 \\
DrQA + Fine-tune & 28.4 & - & 25.7 & - & 19.5 & - & 34.3 & - \\
DrQA + Multitask & 29.8 & - & 25.4 & - & \underline{20.7} & - & 36.5 & - \\
$\text{R}^3$ \cite{wang2017r} & 29.1 & - & 28.4 & - & 17.1 & - & 38.8 & - \\
\midrule
Par. Ranker & 28.5 & 83.1 & 26.8 & 91.4 & 18.0 & 70.7 & 33.4 & 79.7 \\
Par. Ranker + Answer Agg. & 28.9 & - & 28.2 & - & 18.4 & - & 33.9 & - \\
Par. Ranker + Full Agg. & \underline{\textbf{30.2}} & - & \underline{\textbf{35.4}} & - & \textbf{19.9} & - & \underline{\textbf{39.1}} & - \\
\bottomrule
\end{tabular}
\caption{Open-domain QA results on four QA datasets. Best scores including those of the Multitask model are underlined. Bold texts denote best scores excluding those of the Multitask model.}
\label{table:open_result}
\end{table*}

\subsection{Answer Aggregation}
We feed $M$ paragraphs to Document Reader to extract $M$ answers. While Paragraph Ranker increases the probability of including answers in the top $M$ ranked paragraphs, aggregation step should determine the most probable answer among the $M$ extracted answers. Chen et al. \shortcite{chen2017reading} and Clark et al. \shortcite{clark2017simple} used the unnormalized answer probability from the reader. However, as the unnormalized answer probability is very sensitive to noisy answers, Wang et al. \shortcite{wang2017evidence} proposed a more sophisticated aggregation methods such as coverage-based and strength-based re-rankings. \\
\indent In our QA pipeline, we incorporate the coverage-based method by Wang et al. \shortcite{wang2017evidence} with paragraph scores from Paragraph Ranker. Although strength-based answer re-ranking showed good performances on some datasets, it is too complex to efficiently re-rank $M$ answers. Given the $M$ candidate answers $[A_1,...,A_M]$ from each paragraph, we aggregate answers as follows:
\begin{align}
\hat{A} &= \arg\max_{A_i} p(A_i|Q) \notag \\
&= \arg\max_{A_i} \tilde{p}(A_i|P_i, Q)^\alpha p(P_i|Q)^\beta \tilde{p}(D_i|Q)^\gamma
\label{eq:aggregation}
\end{align}
\noindent where $\tilde{p}(A_i|P_i, Q)$ denotes the unnormalized answer probability from a reader given the paragraph $P_i$ and the question $Q$. Importance of each score is determined by the hyperparamters $\alpha$, $\beta$, and $\gamma$. Also, we add up all the probabilities of the duplicate candidate answers for the coverage-based aggregation.

\section{Experiments}
\subsection{Datasets}
We evaluate our pipeline with Paragraph Ranker on the four open-domain QA datasets. Wang et al. \shortcite{wang2017r} termed SQuAD without relevant paragraphs for the open-domain QA as $\textbf{SQuAD}_{\text{OPEN}}$, and we use the same term to denote the open-domain setting SQuAD. \textbf{CuratedTrec} \cite{baudivs2015modeling} was created for TREC open-domain QA tasks. \textbf{WebQuestions} \cite{berant2013semantic} contains questions from Google Suggest API. \textbf{WikiMovies} \cite{miller2016key} contains questions regarding movies collected from OMDb and the MovieLens database. We pretrain Document Reader and Paragraph Ranker on the SQuAD training set.\footnote{On SQuAD development set, pretrained Document Reader achieves 69.1\% EM, and pretrained Paragraph Ranker achieves 96.7\% recall on the top 5 paragraph .}

\subsection{Implementation Details}
Paragraph Ranker uses 3-layer Bi-LSTM networks with 128 hidden units. On $\text{SQuAD}_\text{OPEN}$ and CuratedTrec, we set $\alpha$, $\beta$, and $\gamma$ of Paragraph Ranker to 1. Due to the different characteristics of questions in WebQuestion and WikiMovies, we find $\alpha$, $\beta$, and $\gamma$ based on the validation QA pairs of the two datasets. We use $N=20$ for the number of documents to retrieve and $M=200$ for the number of paragraphs to read for all the four datasets. We use Adamax \cite{kingma2014adam} as the optimization algorithm. Dropout is applied to LSTMs and embeddings with $p=0.4$.


\begin{table*}[t]
\centering
\begin{tabular}{rl}
\toprule
Question \#1 & \textit{What position does Von Miller play?} ($\text{SQuAD}_\text{OPEN}$)\\
Answer & \textit{linebacker}, \textit{linebacker}, \textit{linebacker} \\
\midrule
Doc. Retriever & (Top-1 document) \textit{Ferdinand Miller, from 1875 von Miller ... was an ore caster, ...}\\
 & \textit{Miller was born and died in Munich. He was the son of the artisan and First ...}\\
 & \textit{Ferdinand was simultaneously ennobled. Ferdinand's younger brother was the ...} \\
Par. Ranker & (Top-1 paragraph) \textit{The two teams exchanged field goals ... with a 48-yarder by ... }\\
 & (Top-2 paragraph) \textit{Luck was strip-sacked by Broncos' \underline{\textbf{linebacker}} Von Miller ...} \\
 & (Top-3 paragraph) \textit{Broncos' \underline{\textbf{linebacker}} Von Miller forced a fumble off RGIII ...}\\
\bottomrule
\end{tabular}
\caption{Top ranked paragraphs by Paragraph Ranker based on $\text{SQuAD}_\text{OPEN}$}
\label{table:ranker_analysis}
\end{table*}

\subsection{Results}
In our experiments, \textbf{Paragraph Ranker} ranks only paragraphs, and answers are determined by unnormalized scores of the answers. \textbf{Paragraph Ranker + Answer Agg.} sums up the unnormalized probabilities of duplicate answers (i.e., $\beta=\gamma=0$). \textbf{Paragraph Ranker + Full Agg.} aggregates answers using Equation \ref{eq:aggregation} with the coverage-based aggregation. \\
\indent In Table \ref{table:open_result}, we summarize the performance and recall of each model on open-domain QA datasets. We define recall as the probability of read paragraphs containing answers. While Reinforced Reader-Ranker ($\text{R}^3$) \cite{wang2017r} performs better than DrQA on the three datasets ($\text{SQuAD}_\text{OPEN}$, \text{CuratedTrec}, \text{WikiMovies}), Paragraph Ranker + Full Agg. outperforms both DrQA and $\text{R}^3$. Paragraph Ranker + Full Agg. achieved 3.78\%, 24.65\%, 2.05\%, 0.77\% relative improvements in terms of EM on $\text{SQuAD}_\text{OPEN}$, CuratedTrec, WebQuestion, and WikiMovies, respectively (7.8\% on average). It is noticeable that our pipeline with Paragraph Ranker + Full Agg. greatly outperforms DrQA + Multitask in $\text{SQuAD}_\text{OPEN}$ and \text{CuratedTrec}.

\subsection{Analysis}
In Table \ref{table:ranker_analysis}, we show 3 random paragraphs of the top document returned by Document Retriever, and the top 3 paragraphs ranked by Paragraph Ranker from the top 40 documents. As Document Retriever largely depends on matching of query tokens with document tokens, the top ranked document is usually the document with most tokens matching the query. However, Question 1 includes the polysemy of the word ``play'' which makes it more difficult for Document Retriever to perform effectively. Our Paragraph Ranker well understands that the question is about a sports player not a musician. The top 1-3 paragraphs for the second question came from the 30th, 7th, and 6th documents, respectively, ranked by Document Retriever. This shows that increasing number of documents to rank helps Paragraph Ranker find more relevant paragraphs.

\section{Conclusion}

In this paper, we present an open-domain question answering pipeline and proposed Paragraph Ranker. By using Paragraph Ranker, the QA pipeline benefits from increased answer recall from paragraphs to read, and filters irrelevant documents or paragraphs. With our simple Paragraph Ranker, we achieve state-of-the-art performances on the four open-domain QA datasets with large margins. As future works, we plan to further improve Paragraph Ranker based on the researches on learning to rank.

\section*{Acknowledgement}
This research was supported by National Research Foundation of Korea (NRF-2017R1A2A1A17069645, NRF-2017M3C4A7065887), and the Korean MSIT (Ministry of Science and ICT) under the National Program for Excellence in SW (2015-0-00936) supervised by the IITP (Institute for Information \& communications Technology Promotion)

\bibliography{emnlp2018}
\bibliographystyle{acl_natbib_nourl}

\end{document}